\documentclass[letterpaper]{article} 
\usepackage{aaai23}  
\usepackage{times}  
\usepackage{helvet}  
\usepackage{courier}  
\usepackage[hyphens]{url}  
\usepackage{graphicx} 
\usepackage{natbib}  
\usepackage{caption} 
\usepackage{algorithm}
\usepackage{algorithmic}

\usepackage{newfloat}
\usepackage{listings}
\DeclareCaptionStyle{ruled}{labelfont=normalfont,labelsep=colon,strut=off} 
\floatstyle{ruled}
\newfloat{listing}{tb}{lst}{}
\floatname{listing}{Listing}
\usepackage{amsmath}
\usepackage{graphicx}
\usepackage{amsfonts}
\usepackage{latexsym}
\usepackage{booktabs}
\usepackage{multirow, makecell, caption, multicol}
\usepackage{color,colortbl}
\usepackage[dvipsnames]{xcolor}

\title{Unsupervised Explanation Generation via Correct Instantiations} 
\author{
    Sijie Cheng\textsuperscript{\rm{1,2}}\thanks{\ \ Work done while interning at Shanghai AI Lab.},
    Zhiyong Wu\textsuperscript{\rm2}\thanks{\ \ Corresponding author},
    Jiangjie Chen\textsuperscript{\rm1},
    Zhixing Li\textsuperscript{\rm3}, \\
    Yang Liu\textsuperscript{\rm{4,5}},
    Lingpeng Kong\textsuperscript{\rm{2,6}}
}
\affiliations{
    \textsuperscript{\rm 1}Shanghai Key Laboratory of Data Science, School of Computer Science, Fudan University \\ 
    \textsuperscript{\rm 2}Shanghai AI Laboratory 
    \textsuperscript{\rm 3}Full Truck Alliance \\
    \textsuperscript{\rm 4}Institute for AI Industry Research, Tsinghua University, Beijing, China \\
    \textsuperscript{\rm 5}Department of Computer Science and Technology, Tsinghua University, Beijing, China \\
    \textsuperscript{\rm 6}The University of Hong Kong \\
    \{sjcheng20, jjchen19\}@fudan.edu.cn, wuzhiyong@pjlab.org.cn \\ lizx.2012@tsinghua.org.cn, liuyang2011@tsinghua.edu.cn, lpk@cs.hku.hk 
}

\usepackage{bibentry}

\begin{document}

\maketitle

\begin{abstract}
While large pre-trained language models (PLM) have shown their great skills at solving discriminative tasks, a significant gap remains when compared with humans for explanation-related tasks.
Among them, explaining the reason why a statement is wrong (e.g., against commonsense) is incredibly challenging. 
The major difficulty is finding the conflict point, where the statement contradicts our real world.
This paper proposes \textsc{Neon}, a two-phrase, unsupervised explanation generation framework. \textsc{Neon} first generates corrected instantiations of the statement (\textit{phase I}), then uses them to prompt large PLMs to find the conflict point and complete the explanation (\textit{phase II}). 
We conduct extensive experiments on two standard explanation benchmarks, i.e., ComVE and e-SNLI. 
According to both automatic and human evaluations, \textsc{Neon} outperforms baselines, even for those with human-annotated instantiations. In addition to explaining a negative prediction, we further demonstrate that \textsc{Neon} remains effective when generalizing to different scenarios.
\end{abstract}

\section{Introduction}




Nowadays, Explainable Natural Language Processing (ExNLP) \citep{danilevsky2020survey} has received increasing attention toward trustworthy NLP models. A valid explanation can not only ensure that a model solves a problem using the corresponding knowledge rather than exploiting superficial cues or short-cuts \citep{Niven2019ProbingNN, Geva2019AreWM, Cui2021OnCC}, they can also be used to improve model performance on downstream tasks \citep{Wei2022ChainOT, wang2022self, creswell2022selection}.

\begin{figure}[!t]
    \centering
    \includegraphics[width=1.0\linewidth]{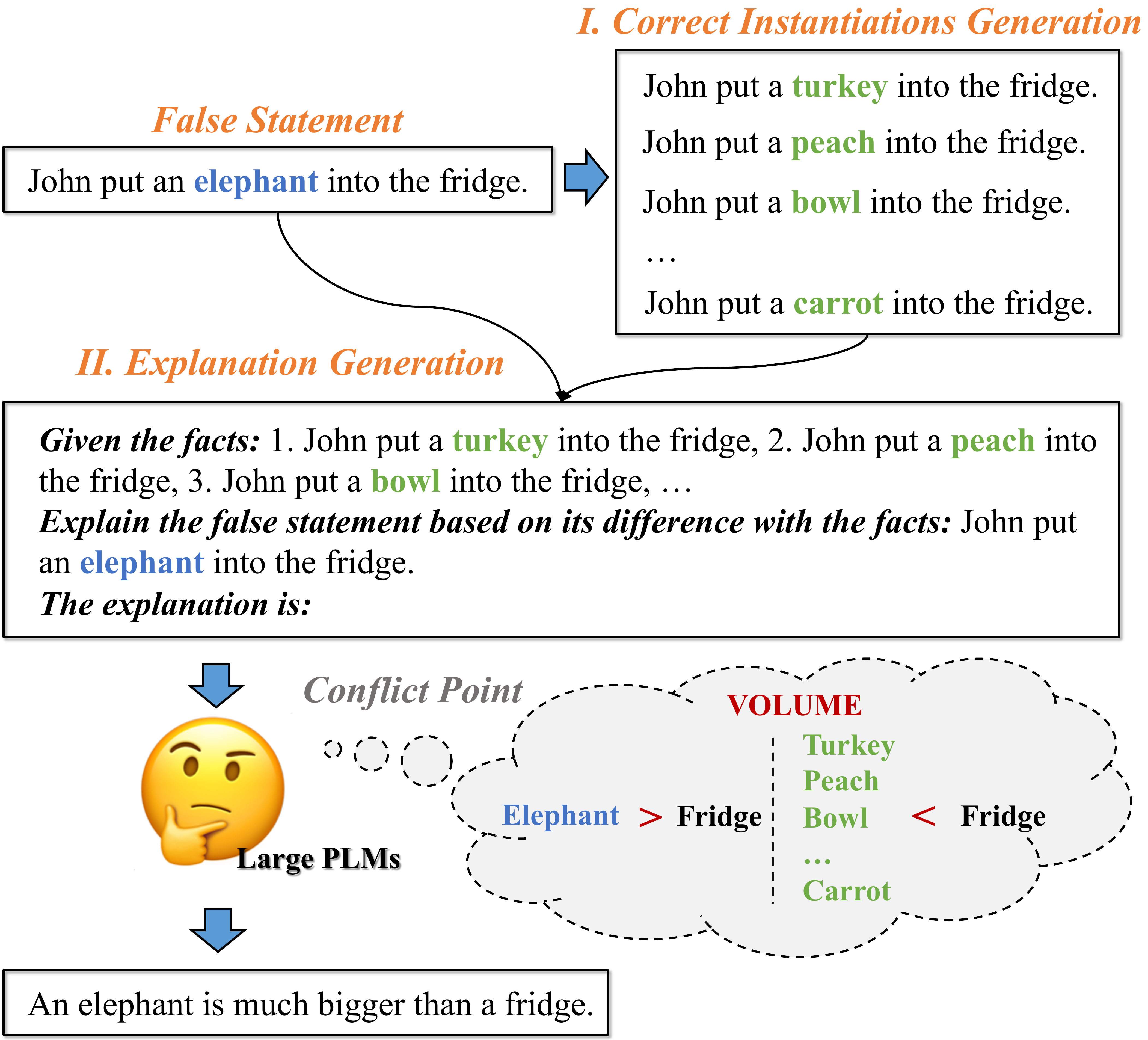}
    \caption{Our proposed two-phase framework \textsc{Neon} (\textit{correct instantiations generation} and \textit{explanation generation}) explains a false statement in ComVE \citep{Wang2020SemEval2020T4} task. The \textit{conflict point} module is implicitly induced inside the large pre-trained language models.}
    \label{fig:intro}
\end{figure}

In general, there are two main types of explanations in the field of ExNLP: highlights and free-text explanations \citep{wiegreffe2021teach}. Highlights \citep{lei2016rationalizing} methods use subsets of the input to support model prediction, thus can not solve the majority of tasks where input does not contain rationales. In this paper, we focus on free-text explanation, which justifies model predictions using natural language. Despite being more flexible and human-readable, free-text explanations pose great challenges~\citep{Lin2020CommonGenAC, rajani2019explain}, as they require models' ability to accurately understand the logic behind the problem and express them in natural language. 

\begin{table*}[t!]
    \centering
    \begin{tabular}{p{6cm}p{8cm}r}
    \toprule
        \textbf{False Statement} & \textbf{Explanation} & \textbf{Conflict Point} \\
        \midrule
        John put an elephant into the fridge. & An elephant is much bigger than a fridge. & Volume \\
        \midrule
        He drinks apple. & Apple can not be drunk. & Function \\
        \midrule
        Jeff ran 100,000 miles today. & No way can someone run 100,000 miles in a day. & Speed \\
        \midrule
        A giraffe is a person. & A giraffe is an animal, not human. & Property \\
        \midrule
        Europe is in France. & Europe is a continent but france is a country. & Geography \\
    \bottomrule
    \end{tabular}
    \caption{Examples and their exact conflict points to explain in ComVE task.}
    \label{tab:example}
\end{table*}

In this paper, we propose a model agnostic framework to generate free-text explanations for \textit{false statements}. Given a false statement which is against commonsense, ``\textit{John put an elephant into the fridge}'', the model is expected to generate a convincing explanation ``\textit{an elephant is much bigger than a fridge}'' to state the reason why the former statement is incorrect~(ComVE; \citealp{Wang2020SemEval2020T4}). Explaining a false statement is generally considered to be more challenging~\citep{wang2019does}, but it is the key to prevent models from making mistakes and to improve their performances \citep{hoffmann2019explainable, lipton1990contrastive}.\footnote{It is worth noting that we also explore explaining correct statements in Section~\ref{sec:generality} to demonstrate the generality of our method.}
Recent studies \citep{Jon2020BUTFITAS, konar2020ana} generally adopt sequence-to-sequence or language model generation approaches. The sequence-to-sequence (seq2seq) methods \citep{Jon2020BUTFITAS, wan2020kalm} use the false statement as the source sequence to generate reason as the target sequence. As for the LM approaches \citep{konar2020ana, Fadel2020JUSTersAS}, they manually design prompts for auto-regressive language models to generate explanations.
We argue that both methods neglect that the key to solve such problems is to find the \textbf{conflict point} as shown in Table \ref{tab:example}, where the false statement contradicts the commonsense knowledge.
For instance, the conflict point of ``\textit{John put an elephant into the fridge}'' is the \underline{relative volume} between ``\textit{elephant}'' and ``\textit{fridge}''.

Finding the exact conflict point can be rather difficult, even for large PLMs.
On the one hand, manually constructing a dataset with conflict points for training is labor-intensive and difficult to scale~\citep{Wang2020SemEval2020T4}.
On the other hand, exact triples of conflict points are rare in the external knowledge graph due to their tacitness and diversity. \citep{wan2020kalm, konar2020ana}.
Considering the limitations of these direct methods mentioned above, we try to provide guided hints as prompts to implicitly elicit PLMs to reason the conflict point, inspired by the line of work about the chain of thought \citep{Wei2022ChainOT, creswell2022selection, wang2022self}. 
To produce guided hints, we automatically generate a bunch of correct instantiations based on the false statement. Then, the conflict points can be implicitly induced from the difference between the commonality of our generated instantiations and the false statement.
For example, given the false statement ``\textit{John put an elephant into the fridge}'', we firstly generate a set of correct instantiations \{``\textit{John put a \textbf{turkey} into the fridge}'', ``\textit{John put a \textbf{peach} into the fridge}'', $\cdots$\} and their underlying commonality is that their volumes are all smaller than the fridge. Combining these instantiations and the false statement, their difference can help PLMs better implicitly reason that the conflict point is relative volume where an elephant is much bigger than a fridge.


In this paper, we propose \textsc{Neon}, a two-phase framework for u\underline{n}supervised \underline{e}xplanation generation via c\underline{o}rrect i\underline{n}stantiations as shown in Figure \ref{fig:intro}.
In the first phase, given the false statement, we attempt both in-context and unsupervised learning to generate correct instantiations automatically. 
In the second phase, combining both generated instantiations and the false statement, the PLMs can implicitly induce the conflict point better to generate explanations. To purely detect the ability of implicit induction in off-the-shelf PLMs, we explore the model performance in the unsupervised setting rather than the traditional supervised setup \citep{Wang2020SemEval2020T4}. We conduct extensive experiments on two standard explanation benchmarks, ComVE \citep{Wang2020SemEval2020T4} and e-SNLI \citep{camburu2018snli}. Experimental results prove the effectiveness of our method on both automatic and manual evaluations. Furthermore, we also conduct analysis experiments to demonstrate that the main idea of \textsc{Neon} can generally be extended to accommodate other explanation tasks.

The contributions of our work are as follows:
\begin{itemize}
    \item We propose a novel method based on the importance of conflict points to solve the false statement explanation problem. To the best of our knowledge, we are the first to introduce the concept of the conflict point in the task.
    \item We propose a two-phase framework named \textsc{Neon} to elicit the large PLMs to induce through instantiations to unsupervisedly generate explanations.
    \item We present analyses of our generated instantiations and demonstrate the generality of \textsc{Neon}.
\end{itemize}


\section{Methodology}

\subsection{Problem Formulation}
Our target problem is to generate a reason to explain why the false statement does not make sense or is inconsistent. Given the original false statement with $n$ tokens $\boldsymbol{x} = \{x^1, x^2, \cdots, x^n\}$, we automatically generate a set of correct instantiations $\mathbb{H} = \{\boldsymbol{h}_1, \boldsymbol{h}_2, \cdots, \boldsymbol{h}_l\}$ with a commonality.
Each instantiation $\boldsymbol{h} = \{h^1, h^2, \cdots, h^k\}$ with $k$ tokens is a constrained text generating conditioned on the original false statement $\boldsymbol{x}$.
According to both the false statement $\boldsymbol{x}$ and our generated correct instantiations $\mathbb{H}$, the pre-trained language model $G$ needs to implicitly reason the conflict point and give a rational explanation with $m$ tokens $\boldsymbol{y} = \{y^1, y^2, \cdots, y^m\}$.

\begin{table}[t!]
    \centering
    \begin{tabular}{c}
    \toprule
    \rowcolor[gray]{0.95} 
    \textbf{Phase I: Correct Instantiations Generation} \\
    \midrule
    \makecell[l{p{8cm}}]{Task: Based on the incorrect statement, generate the correct statement. \\
    \\
    \textcolor{gray}{/* Example 1 */} \\
    Incorrect statement: \textit{He drinks apple}. \\
    Correct statement: \textbf{\textit{He drinks milk.}} \\
    \\
    \textcolor{gray}{/* Test data */} \\
    Incorrect statement: \textit{John put an elephant into the fridge.} \\
    Correct statement:} \\
    \midrule
    \rowcolor[gray]{0.95} 
    \textbf{Phase II: Unsupervised Explanation Generation} \\
    \midrule
    \makecell[l{p{8cm}}]{Given the facts: \textbf{\textit{1. John put a turkey into the fridge, 2. John put a peach into the fridge, 3. John put a bowl into the fridge,}} \\
    Explain the following statement based on its difference with the facts: \textit{John put an elephant into the fridge}.\\
    The explanation is:}  \\
    \bottomrule
    \end{tabular}
    \caption{The prompt instances of in-context learning in our two phases: presented are the \textit{incorrect statements} and the \textbf{\textit{correct statements}}. We use 16 examples per prompt in the first phase. More details can be found in Appendix \ref{app:Prompt}.}
    \label{tab:incontext}
\end{table}

\subsection{Correct Instantiations Generation}
In the first phase, we attempt two different means to generate correct instantiations $\mathbb{H}$ conditioned on the false statements $\boldsymbol{x}$ to prove the flexibility of our framework \textsc{Neon}.
One adopts in-context learning with larger language models under the few-shot setting, the other one is based on traditional constrained text generation in an unsupervised way.

\paragraph{In-context Learning}
Considering the large number of parameters in PLMs, in-context learning uses a series of demonstrations of a specific task as prompts to induce model generation, while the parameters of large PLMs are fixed \citep{brown2020language, radford2019language}. Besides the advantage of no extra need to train or finetune, in-context learning can also reduce the reliance on large amounts of annotated data.
Therefore, due to the great performance of in-context learning with large PLMs in the recent studies \citep{wiegreffe2021reframing}, we attempt to use in-context learning to generate correct instantiations $\mathbb{H}$ automatically given the original false statement $\boldsymbol{x}_\textnormal{ori}$.

Following \citet{wiegreffe2021reframing}, we apply the in-context learning under few-shot setups to generate our correct instantiations. We specifically design a prompt followed by the false statement that the model needs to correct. To construct the prompt, we firstly randomly sample 200 instances ([correct statement, incorrect statement] in the ComVE task and [entailment statement, contradiction statement] in the e-SNLI task) from the training dataset. Then we randomly select $K$ instances and concatenate them to construct our prompt as shown at the top of Table \ref{tab:incontext}. Finally, we feed the model with both constructed prompt and our test data to infer the completion which can be regarded as our generated correct instantiations.


\paragraph{Constrained Text Generation}
Despite its simplicity, in-context learning often requires human annotations, which is not always available. In this section, we explore the challenging setting where instantiations are generated in a fully unsupervised manner.
As a preliminary study, we apply the widely used constrained text generation framework CGMH~\citep{miao2019cgmh}.

Considering that our first step is to detect the conflicting position, we adopt perplexity (PPL) to evaluate the influence one token has on the whole statement. The perplexity score is computed by the masked language models, such as RoBERTa \citep{liu2019roberta}.
Given the false statement $\boldsymbol{x} = \{x^1, x^2, \cdots, x^n\}$, we compute the relatively perplexity score of the original statement to its masking sentence which replaces $x^i$ with \texttt{[MASK]}. 
If the $i$-th token is unlikely to exist in the position, the perplexity score $S_\textnormal{PPL}^i$ is larger, which indicates the token should be edited with a higher priority.

\begin{equation}
    S_\textnormal{PPL}^{i} = \frac{\textnormal{PPL}(\boldsymbol{x})}{\textnormal{PPL}(\boldsymbol{x}\backslash\{x^i\})}
\end{equation}




Given the sampled positions, we need to determine each position's action. Our token-level actions mainly include three types following \citet{chen2021unsupervised}: \textit{insert}, \textit{delete} and \textit{replace}. More details can be found in Appendix \ref{app:cgmh}.
As for the acceptance rate to each generated sentence $\boldsymbol{x}'$ with edited action, our considering property is \textbf{fluency} which is important to guarantee in generative tasks. We measure this fluency score via computing likelihood based on the auto-regression language models, e.g., GPT-2 \citep{radford2019language}.

\begin{equation}
    S_{\textnormal{Fluency}} = \prod_{i=1}^{n}P_{\textnormal{LM}}(h^i|h^{<i})
\end{equation}



\begin{table*}[t!]
    \centering
    \resizebox{\textwidth}{22mm}{
    \begin{tabular}{c|l|cccc|cccc}
        \toprule
        \multirow{2}{*}{\textbf{Row}} & \multirow{2}{*}{\textbf{Method}} & \multicolumn{4}{c|}{\textbf{ComVE}} & \multicolumn{4}{c}{\textbf{e-SNLI}}  \\
        \cmidrule(lr){3-6} \cmidrule(lr){7-10}
        & & \textbf{BLEU} & \textbf{ROUGE} & \textbf{BERTScore} & \textbf{S-BERT} & \textbf{BLEU} & \textbf{ROUGE} & \textbf{BERTScore} & \textbf{S-BERT}\\
        \midrule
        1 & \textbf{Random} & 1.47 & 17.81 & 46.21 & 42.54 & 4.94 & 24.23 & 50.73 & 43.05 \\
        2 & \textbf{Retrieval-BM25} & 1.51 & 17.23 & 45.18 & 38.68 & 4.29 & 23.31 & 49.80 & 42.09 \\
        3 & \textbf{Retrieval-SBERT} & 1.69 & 18.55 & 46.64 & 45.47 & 4.64 & 24.45 & 51.16 & 48.22 \\
       4 &  \textbf{Original} & 1.88 & 20.21 & 48.68 & 51.82 & 4.71 & 25.38 & 50.92 & 46.39 \\
        5 & \textbf{Ground-truth} & 2.48 & 21.25 & 49.66 & \textbf{55.21} & 5.57 & 25.62 & 51.96 & 49.19 \\
        6 & \textbf{Top-1} & 2.42 & 21.42 & 49.86 & 55.03 & 6.03 & 25.87 & 51.97 & 48.51 \\
        7 & \textsc{\textbf{Neon}} \textbf{w/ CGMH} & 3.37 & 20.10 & 48.92 & 49.50 & 4.67 & 26.04 & 51.04 & 48.42 \\
        8 & \textsc{\textbf{Neon}}  \textbf{w/ In-context} & \textbf{3.39} & \textbf{22.50} & \textbf{51.50} & 54.52 & \textbf{6.20} & \textbf{27.28} & \textbf{53.87} & \textbf{51.69} \\
        \bottomrule
    \end{tabular}}
    \caption{The automatic evaluation results of ComVE and e-SNLI tasks.}
    \label{tab:results}
\end{table*}

\subsection{Unsupervised Explanation Generation}

As demonstrated in recent studies \citep{zhong2022describing}, PLMs can capture subtle yet critical differences between different groups of sentences. This inspires us that capturing the differences between the false statement $\boldsymbol{x}$ and our generated instantiations $\mathbb{H}$ can help PLMs induce conflict points.
Therefore, in the second phase, given both the false statement $\boldsymbol{x}$ and our generated instantiations $\mathbb{H}$, we implicitly induce the large PLMs to generate the free-text explanation $\boldsymbol{y}$ in the zero-shot setting. 

\paragraph{Zero-shot Learning}

We adopt a similar prompt construction strategy as discussed in the correct instantiations generation phase. However, unlike the template of few-shot learning in instruction style, our template of zero-shot learning is more fluency like a complete sentence, following previous studies \citep{sanh2021multitask}. In particular, we directly design a natural language description according to different tasks instead of sampled exemplars from training datasets as shown at the bottom of Table \ref{tab:incontext}. Considering the variance due to different descriptions, more analysis on the design of prompting can be found in Section \ref{sec:number} and \ref{sec:robust}. Finally, based on our constructed prompt, the PLMs generate the completion as our generated explanation.

\section{Experiments}

\subsection{Experimental Setups}
\paragraph{Datasets} 
Our experiments are conducted on the two important explanation benchmarks, ComVE \citep{Wang2020SemEval2020T4} and e-SNLI \citep{camburu2018snli}. The ComVE task asks annotators to create 11,997 instances in the format of $\{c_n, s_n, r_{n_1}, r_{n_2}, r_{n_3}\}$, where $c_n$ and $s_n$ are the correct and incorrect statement, respectively. $\{r_1, r_2, r_3\}$ are three reference reasons to explain the incorrect statement. Then they divide all these annotated instances into train/dev/test datasets with 10,000/997/1,000 instances. As for the e-SNLI task, the $c_n$ and $s_n$ can be seen as entailment and contradiction statements, respectively. Filtering the odd instances with only entailment or contradiction statement, our obtained train/dev/test is 5,189/3,280/2,640.


\paragraph{Models} 
In our main experiments, We all adopt the large pre-trained language model OPT-175B \citep{zhang2022opt}. 
To ensure the generalization of our framework, we also conduct experiments on other PLMs varying from small model scale to large. More details can be found in Section \ref{sec:model}.

\paragraph{Implementation Details} 
To fix the max-length of the context window ($n_{ctx}=2048$), we set the number of examples in prompt as $K=16$ in the first phase.
In the first phase, the max length of generated instantiations is 25 for ComVE and 40 for e-SNLI. As for the constrained text generation, we adopt GPT-2-large and RoBERTa-large. In the second phase, the max length of generated explanations is 30 for both tasks. We set the hyper-parameter of Top-p as 0.9, and temperature as 0 for all generation models. We repeat the same experiment three times and report the average accuracy for all experiments. Our experiments are conducted with 8 A100 GPUs.

\paragraph{Baselines}
We compare our framework with the following baselines. (1) \textbf{Original}: the model only feeds with the false (contradiction) statement $s_n$ to generate its rational explanation. (2) \textbf{Random}: We feed a randomly sampled human-annotated correct (entailment) statement and the false statement into the model. (3) \textbf{Retrieval}: we adopt both BM25 \citep{robertson2009probabilistic} and Sentence-BERT (SBERT) \citep{reimers2019sentence} to retrieve the five nearest statements from the Open Mind Common Sense (OMCS) corpus \citep{singh2002open}, then give them and the false statement into the model. (4) \textbf{Ground-truth}: we offer both the false statement $s_n$ and its corresponding human-annotated correct statement $c_n$ to the model. It is worth noting that the human-annotated statement can be regarded as the upper bound of our generated single instantiation. (5) \textbf{
Top-1}: the model generates explanations based on the false statement and our Top-1 generated correct instantiation. To ensure fairness, we keep the templates of all these baselines (except for the original baseline) the same as ours.

\subsection{Evaluation Metrics} \label{sec:metrics}

\paragraph{Automatic Evaluation Metrics}
Considering that the official automatic evaluation metric BLEU \citep{papineni2002bleu} is too harsh to evaluate the quality of explanations \citep{zhao2019moverscore, konar2020ana, Fadel2020JUSTersAS}, we further involve a set of common evaluation metrics as supplementary following \citet{becker2021reconstructing}. In detail, we measure diverse aspects, including token overlap using BLEU and ROUGE \citep{lin2004rouge}, semantic similarity using both BERTScore \citep{zhang2019bertscore} and S-BERT \citep{reimers2019sentence}.

\paragraph{Manual Evaluation Metrics}
Due to the limitation of existing automatic metrics in the open-ended text generation community \citep{zhang2019bertscore, novikova2017we}, we further conduct manual evaluations to compensate following \citet{wiegreffe2021reframing}.
Firstly, we randomly select 100 samples from the test set. We then evaluate generated explanations through head-to-head comparisons.
In order to directly reflect the impact of our instantiations, three annotators are asked to choose the better explanation between the original baseline and \textsc{Neon}. 
To ensure fairness, we shuffle all the generated explanations to be unordered.
We specifically design two aspects: one is the \textit{preferred explanation} from the comprehensive consideration, and the other one needs to express the \textit{conflict points} explicitly. 

\subsection{Results}
\paragraph{Automatic Evaluation}
Table~\ref{tab:results} shows the automatic evaluation results of \textsc{Neon} and baselines. To illustrate the effectiveness of introducing instantiations, we first compare the Original baseline (without instantiations) against others (Row 5-8 in Table~\ref{tab:results}). As we can see, incorporating instantiations in explanation generation consistently improves model performance over the baseline without instantiations. Given the necessity of instantiations, we further investigate how the quality of instantiations affects performance. We observe significant performance deterioration when equipping the model with instantiations that come from random knowledge (Row 1) or strong retrieval models (Row 2-3). This indicates that introducing irrelevant or inaccurate information would hurt model performance. The above comparison also verifies the effectiveness of our instantiation generation method, which is further supported by the comparable performance between the Top-1 and Ground-truth baseline. Finally, by comparing \textsc{Neon} with Top-1 and Ground-truth, we find that ensemble multiple instantiations that share a commonality significantly outperforms baselines on almost all metrics in both ComVE and e-SNLI tasks. We hypothesize that ensembling similar instantiations can help the model better locate the conflict points. This hypothesis is later supported by both human evaluation results (see below) and nuanced analysis of instantiations (see Section~\ref{sec:quality}).

Despite the excellent performance of \textsc{Neon} under the in-context setting, we find it barely outperforms the Original baseline when editing is performed in a fully unsupervised manner (Table~\ref{tab:results}, Row 7). The reason could be that we use relatively small PLMs (Roberta-large and GPT-large) in CGMH due to computation constraints, whereas we use OPT-175B for in-context editing. We leave it as future work to investigate how to apply CGMH on huge PLMs for instantiation generation. Given the performance gap, all later analyses will be based on \textsc{Neon} w/ In-context. 

\paragraph{Manual Evaluation}
As shown in Table \ref{tab:head2head}, for both ComVE and e-SNLI tasks, \textsc{Neon} outperforms the original baseline with respect to \textit{preferred explanation} and \textit{conflict point}. The proportional distribution of the preferred explanation is similar to conflict point, which supports our claim that it is important to find conflict point to generate good explanations. In the conflict point aspect, the fact that \textsc{Neon} beat the original baseline reflects the contribution of our generated instantiations.
It is worth noticing that there still remains a significant proportion of ties (40\%). We believe a better method of finding conflict points can contribute to closing this gap.

\begin{table}[htbp!]
    \centering
    \begin{tabular}{lcccc}
    \toprule
        \multirow{2}{*}{\textbf{Dataset}}& \multicolumn{3}{c}{\textbf{Preferred Explanation (\%)}} & \multirow{2}{*}{$\kappa$}\\
        \cmidrule(lr){2-4}
        & \textbf{Original} & \textbf{Tie} & \textbf{\textsc{Neon}} &  \\
        \midrule
        \textbf{ComVE} & 20.33 & 42.67 & 37.00 & 0.47 \\
        \textbf{e-SNLI} & 18.67 & 41.67 & 39.67 & 0.39 \\
        \midrule
        & \multicolumn{3}{c}{\textbf{Conflict Point (\%)}} \\
        \cmidrule(lr){2-4}
        \textbf{ComVE} & 19.33 & 46.00 & 34.67 & 0.45 \\
        \textbf{e-SNLI} & 15.67 & 53.67 & 30.67 & 0.36 \\
     \bottomrule
    \end{tabular}
    \caption{Head-to-head human evaluation for 100 explanations generated by the original baseline and \textsc{Neon}. Results are shown as \% preferences with Fleiss Kappa $\kappa$.}
    \label{tab:head2head}
\end{table}



\section{Analysis}


\subsection{Quality of Generated Instantiations} \label{sec:quality}

\paragraph{Automatic Evaluation}
To check the correctness of generated instantiations, we fine-tune RoBERTa-Large \citep{liu2019roberta} on both training datasets as binary classifiers. It achieves the accuracy of 88.97 and 84.25 on the ComVE and e-SNLI, respectively. We use these fine-tuned RoBERTa models to evaluate the quality of our generated instantiations. Because the performance of in-context learning is much better than CGMH in our first phase. We conduct experiments mainly on in-context learning in our analyses.

\paragraph{Manual Evaluation} 
Following previous studies \citep{wiegreffe2021reframing}, we assume that the desired instantiations need to meet the requirements at least in terms of both surface and explanation levels. Therefore, we further conduct manual evaluations to assess our generated instantiations through five primary criteria: \romannumeral1. \textit{Acceptability} - Generated instantiations are acceptable in overall judgment;  \romannumeral2. \textit{Grammaticality} - Generated instantiations should be at least fluent with no grammatical mistakes; \romannumeral3. \textit{Factuality} - Generated instantiations should be factually correct; \romannumeral4. \textit{Diversity} - We expect to generate more diverse instantiations; \romannumeral5. \textit{Commonality} - Generated instantiations are expected to have a commonality to help large PLMs infer the conflict point. We randomly select 100 samples from the test set and their corresponding generated instantiations. Then, after shuffling all selected samples, three annotators are asked to choose acceptable/unacceptable for the acceptability metric and use a 3-point Likert-scale rating to evaluate sampled data for the other four aspects.

\paragraph{Results}
We evaluate the quality of the automatic generated and human generated instantiations on ComVE and e-SNLI, they reached the accuracy of 70.28/89.60 and 92.30/97.84, respectively.
Note that in-context learning only uses a few exemplars in the prompts.
As shown in Table \ref{tab:quality}, the human acceptance of the generated instantiations is 72.80/81.67, consistent with the results of the automatic evaluation discussed above. 
As for the surface-level criteria, the score of grammaticality is pretty high, while the score of factuality is relatively worse. The results of the diversity and commonality metrics are over 2.5 points, indicating that the instantiations have a high diversity while sharing a common underlying property well.

\begin{table}[htbp!]
    \centering
    \small
    \begin{tabular}{cccccc}
    \toprule
        \textbf{Dataset} & \textbf{Acc.} & \textbf{Gram.} & \textbf{Fact.} & \textbf{Diver.} & \textbf{Common.} 
        \\
        \midrule
        \textbf{ComVE} & 72.80 & 2.97 & 2.66 & 2.63 & 2.56 \\
        \textbf{e-SNLI} & 81.67 & 2.88 & 2.72 & 2.89 & 2.66 \\
    \bottomrule
    \end{tabular}
    \caption{The manual evaluation results of our generated instantiations. 
    (\textbf{\romannumeral1.} Acceptability \textbf{\romannumeral2.} Grammaticality \textbf{\romannumeral3.} Factuality \textbf{\romannumeral4.} Diversity \textbf{\romannumeral5.} Commonality)}
    \label{tab:quality}
\end{table}

\begin{table}[h!]
    \small
    \centering
    \begin{tabular}{lcccc}
    \toprule
        \textbf{Method} & \textbf{BLEU} & \textbf{ROUGE} & \textbf{BERTScore} & \textbf{S-BERT} \\
        \midrule
        \textbf{Top-1} & \textbf{2.47} & 20.77 & 49.13 & 54.25 \\
        \textbf{Top-1*} & 2.20 & \textbf{21.39} & \textbf{49.63} & \textbf{54.98} \\
        \midrule
        \textbf{\textsc{Neon}} & 3.39 & 21.65 & 49.09 & 53.11 \\
        \textbf{\textsc{Neon*}} & \textbf{3.51} & \textbf{22.32} & \textbf{49.54} & \textbf{54.53} \\
    \bottomrule
    \end{tabular}
    \caption{The comparison of our methods before (marked *) and after filtering the low-quality instantiations.}
    \label{tab:filter}
\end{table}

Furthermore, we filter the low-quality instantiations determined by the automatic metric to probe the correlation between the quality and model performance. Taking the ComVE task as an example, we firstly generate 10 instantiations per data for 1,000 test data and filter low-quality instantiations. We then obtain 987 and 773 samples with Top-1 and ensemble (five) high-quality instantiations, respectively. Finally, we compare the model performance before and after filtering. The results are shown in Table \ref{tab:filter}. As for the Top-1 instantiation, the results before and after filtering are comparable which supports the similar situation between our ground-truth and Top-1 baselines reported in Table \ref{tab:results}. Our ensemble method shows a general improvement in model performance after filtering. Combining the above phenomena, we believe that the ensemble method is more stringent about the quality of generated instantiations due to asking for commonality among them.


\begin{table}[t!]
    \centering
    \begin{tabular}{c|cccc}
    \toprule
        \textbf{\#} & \textbf{BLEU} & \textbf{ROUGE} & \textbf{BERTScore} & \textbf{S-BERT} \\
        \midrule
        \textbf{1} & 2.42 & 21.03 & 49.22 & 52.70 \\
        \textbf{2} & 2.61 & 21.14 & 49.22 & 52.56 \\
        \textbf{3} & 3.32 & 21.32 & 49.46 & 51.79 \\
        \textbf{4} & 3.29 & 22.26 & 50.97 & \textbf{54.74} \\
        \textbf{5} & 3.39 & \textbf{22.50} & \textbf{51.50} & 54.52 \\
        \textbf{6} & 3.01 & 21.49 & 49.11 & 49.06 \\
        \textbf{7} & \textbf{3.48} & 21.57 & 49.45 & 49.66 \\
        \textbf{8} & 3.28 & 21.27 & 49.66 & 49.94 \\
        \textbf{9} & 3.16 & 21.70 & 49.91 & 48.73 \\
        \textbf{10} & 3.39 & 21.21 & 49.94 & 49.47 \\
    \bottomrule
    \end{tabular}
    \caption{Model performance with the different number of ensemble instantiations in the ComVE task.}
    \label{tab:number}
\end{table}

\subsection{Effects on Instantiations Number} \label{sec:number}

Despite the fact that the PLMs can implicitly induce the conflict points through instantiations, it still remains a question of whether more generated instantiations lead to better performance. Therefore, we detect the model performance with the different numbers of generated instantiations varying from 1 to 10. The results are shown in Table \ref{tab:number}.\footnote{To keep the templates consistent, we separate each instance by an ordinal number, including only one instantiation.}
When the number of instantiations increases from 1 to 5, the model performance exhibits an upward trend. This phenomenon indicates that the increasing diversity of generated instantiations decreases the possibility of other misleading conflict points.
However, as the number of instantiations increases from 6 to 10, the model performance plateaus.
We conjecture there are two-fold reasons. One is that sufficient diversity and more noise will limit the improvement of model performances when the number reaches a certain level.
The other one is that prompts containing overlong and unoriginal sequences will damage the performance.

\subsection{Effects on Model Size} \label{sec:model}
\begin{figure}[t!]
    \centering
    \includegraphics[width=1.0\linewidth]{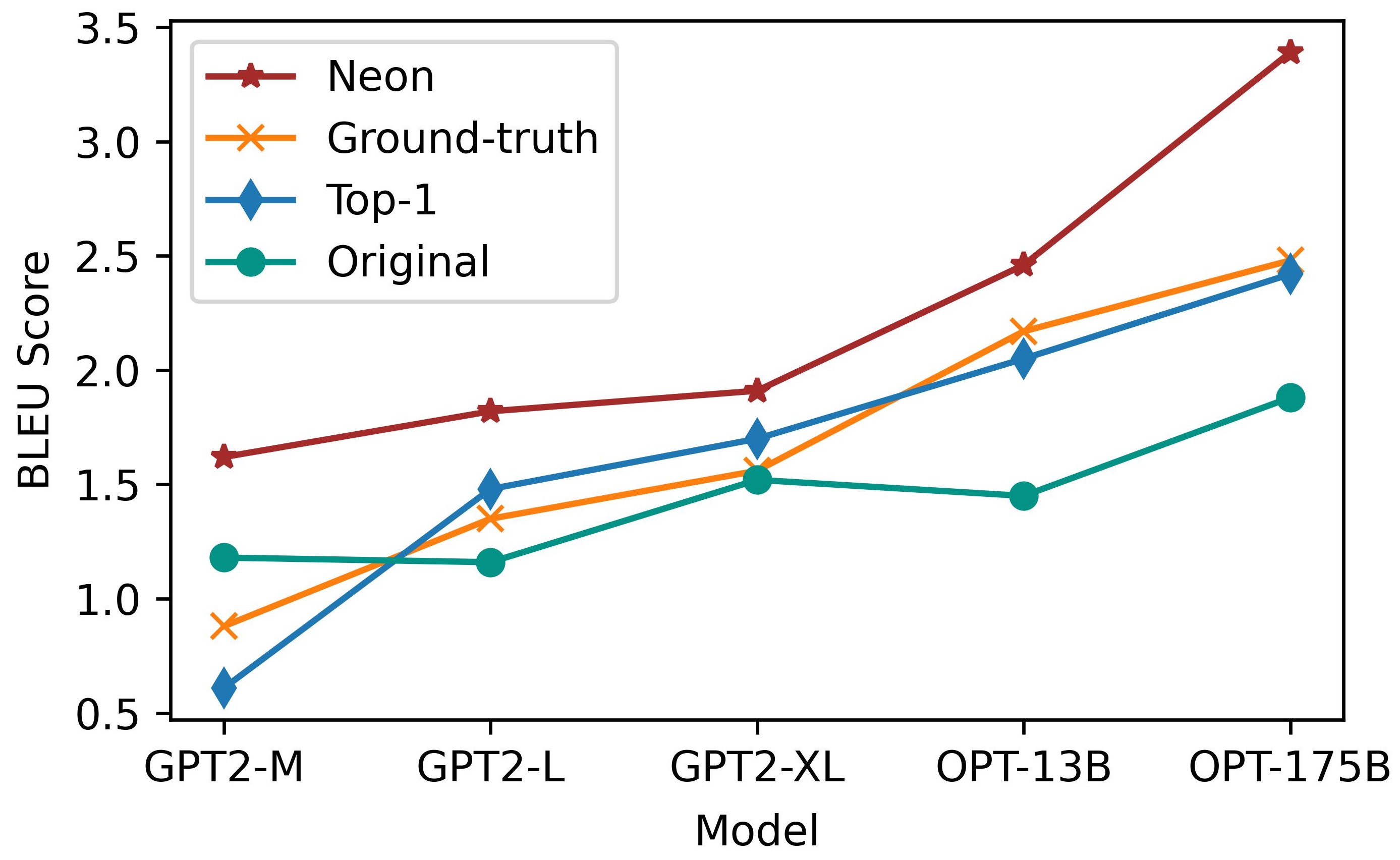}
    \caption{Model performance of increasing model scales in the ComVE task. More results can be found in Appendix \ref{app:model}.}
    \label{fig:model_size}
\end{figure}

In this section, we detect the model performance to generate explanations with increasing model scales. As shown in Figure \ref{fig:model_size}, the experimental results are similar in most of these models, except for the smallest model GPT2-M.
This phenomenon indicates that only offering an extra instantiation will be regarded as noise to hurt model performances when the model parameter is relatively small.
However, our proposed method with ensemble instantiations obviously beat all other baselines with different model scales, reflecting its robustness.
Moreover, as the scale of model parameters increases, the performance gap between the proposed method and the baselines becomes larger. This trend shows that the implicit induction through instantiations of large PLMs is an emerging ability with increasing model scales, which is consistent with previous studies \cite{Wei2022ChainOT, wang2022self}.

\subsection{Robustness of Prompting} \label{sec:robust}
According to previous studies \citep{zhao2021calibrate, lu2021fantastically}, the model performances are sensitive to templates.
Therefore, we further evaluate the robustness of \textsc{Neon} following \citet{Wei2022ChainOT}. Another two annotators B and C are asked to write the templates independently. Furthermore, we ask annotator A to write an instruction-style template that is more concise than before, following \citet{cobbe2021training}. The written templates are shown in Appendix \ref{app:robust}. Results shown in Figure \ref{fig:robust} indicate that though there exists a variance among different annotated templates, all our prompts still outperform the original baseline. However, the instruction style prompt is significantly worse than the natural language description style in the zero-shot setting, due to the lack of instruction style signals in the pre-training corpus. 

\begin{figure}[h!]
    \centering
    \includegraphics[width=1.0\linewidth]{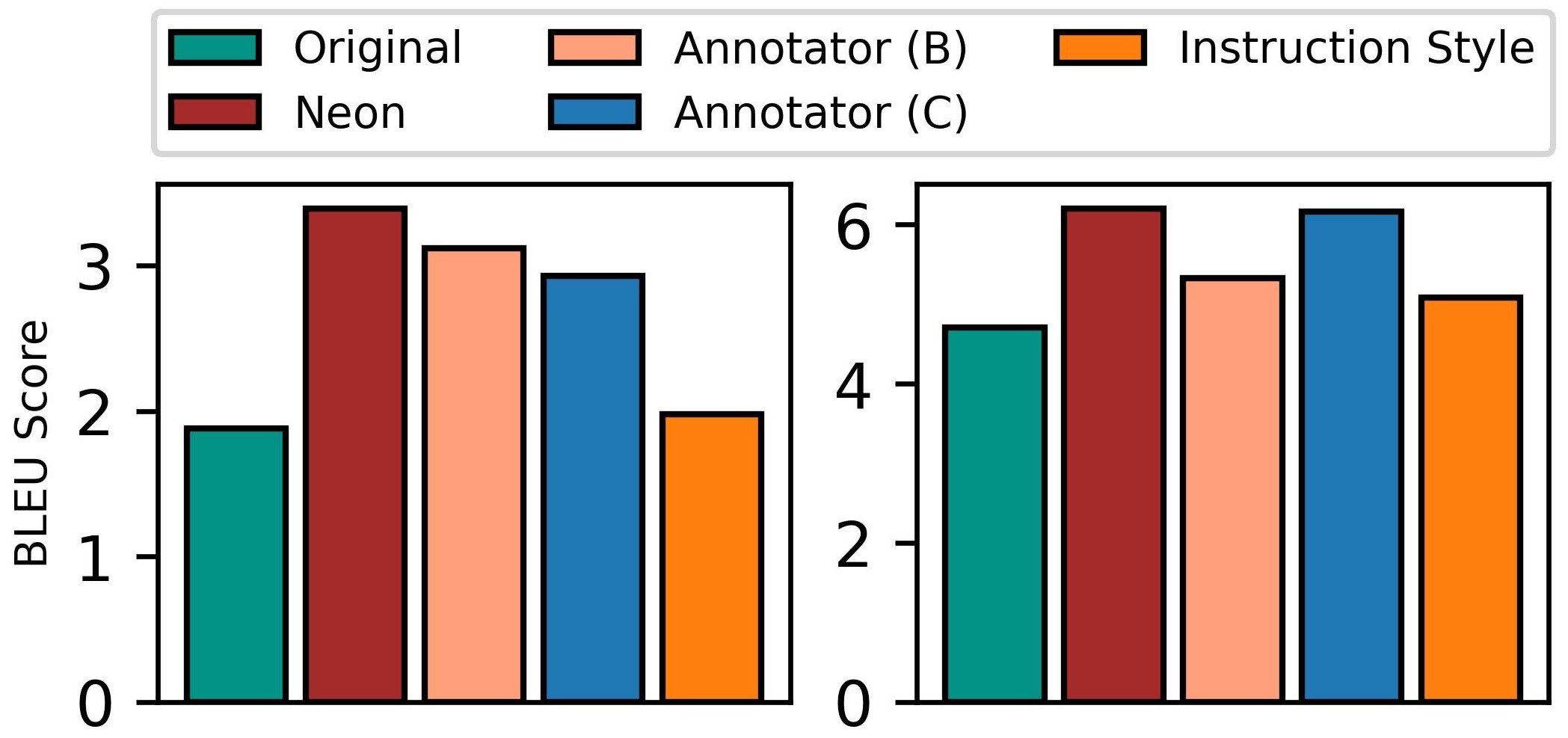}
    \caption{Independently-written for the robustness of \textsc{Neon}. Results of other metrics can be found in Appendix \ref{app:robust}.}
    \label{fig:robust}
\end{figure}

\subsection{Demonstration of Generality} \label{sec:generality}

In this section, we adapt \textsc{Neon} to generate explanations for correct statements. Taking the e-SNLI task as an example, given the entailment statement $c_n$, there are three human-annotated explanations $\{r'_1, r'_2, r'_3\}$. We directly use the generated correct instantiations in the first phase as guided hints. We find that their commonality with the entailment statement can help large PLMs to explain. The results are shown in Table \ref{tab:generality}. Notably, the top-1 baseline is slightly worse than the original baseline. This phenomenon suggests that a single instantiation no longer provides valid information like the contrast in false statements during the explanation of correct statements. 
However, \textsc{Neon} still significantly outperforms other baselines, which demonstrates the effectiveness and generality.

\begin{table}[htbp!]
    \small
    \centering
    \begin{tabular}{ccccc}
    \toprule
        \textbf{Method} & \textbf{BLEU} & \textbf{ROUGE} & \textbf{BERTScore} & \textbf{S-BERT} \\
         \midrule
        \textbf{Original} & 8.11 & 29.73 & 52.66 & 53.18 \\
        \textbf{Top-1} & 9.22 & 28.64 & 52.64 & 50.81 \\
        \textbf{\textsc{Neon}} & \textbf{11.18} & \textbf{31.69} & \textbf{55.30} & \textbf{56.33} \\
    \bottomrule
    \end{tabular}
    \caption{Model performance of generating explanations for correct statements in the e-SNLI task.}
    \label{tab:generality}
\end{table}

\subsection{Case Study}
Finally, we present some generated examples of two phases as shown in Table \ref{tab:case_study}. In the first phase, the quality of our generated instantiations is comparable to human-annotated instantiations. Especially, the keywords of ``safety'', ``shelter'' and ``peace'' meet the needs of both diversity and commonality.
As for the second phase, \textsc{Neon} and other baselines yield fluent explanations. However, the original baseline does not give a correct definition of \emph{home}, and the ground-truth baseline pays much attention to the keyword ``security'' so that its generated explanation is irrelevant. Overall, \textsc{Neon} products a better explanation thanks to the commonality induced from the generated instantiations in the first phase.

\section{Related Work}
\paragraph{Natural Language Explanations}
To improve the transparency and reliability of large deep learning language models, a promising approach is to ask models to generate natural language free-form explanations. 
This explanation technique is first introduced in computer vision applications \citep{park2018multimodal, hendricks2016generating}.
Then it broadcasts into diverse domains of natural language processing. 
For instance, \citet{camburu2018snli} annotate natural language explanations of the entailment relations in the Stanford Natural Language Inference dataset. \citet{rajani2019explain, Wang2020SemEval2020T4, aggarwal2021explanations} extends this idea into commonsense explanations. 
To solve these explanation tasks, traditional methods \citep{Jon2020BUTFITAS, konar2020ana} usually finetune generation models with explicit human supervision.
\citet{wan2020kalm, konar2020ana} exploit external knowledge graph to fill in the needed background in the explanation.
Most relevant to our study is the contrastive explanation \citep{paranjape2021prompting, ross2020explaining}, which gives a contrastive explanation to answer to ``Why P rather than Q''. However, they often focus on only one fact. In contrast, we notice that ensemble instantiations with a commonality can help find the exact conflict point.

\paragraph{In-context Learning}
After the fine-tuning paradigm of large PLMs, in-context learning has been attractive due to its simple operation and strong interaction. More important, it does not have to update the model parameters anymore. \citet{brown2020language} propose that large PLMs can complete a generation given a few demonstrations as prompts.
Recently, more studies have paid attention to generating rationales through in-context learning to help language model performance. \citet{Wei2022ChainOT} adopt ``chain-of-thought'' reasoning prompt to induce large PLMs reason step-by-step. Similarly, \citep{wang2022self} explore that ensemble rationales can significantly improve the model performance. \citep{zelikman2022star} propose a bootstrapping strategy to improve the quality of rationale examples.

\begin{table}[t!]
    \centering
    \small
    \begin{tabular}{c}
    \toprule
        \rowcolor[gray]{0.95} 
        \textbf{Phase I: Correct Instantiations Generation} \\
        \midrule
        \makecell[l{p{8cm}}]{
        \textbf{False Statement:} A home is a place for violence.\\
        \textbf{\textsc{Neon}:} 1. A home is a place for safety. 2. A home is a place for love. 3. A home is where you live. 4. A home is a place for shelter. 5. A home is a place of peace.  \\
        \textbf{Human-annotated:} A home gives a person a sense of security. \\
        }
        \\
        \midrule
        \rowcolor[gray]{0.95} 
        \textbf{Phase II: Unsupervised Explanation Generation} \\
        \midrule
        \makecell[l{p{8cm}}]{
        \textbf{Original:} That's the wrong definition of home. A place isn't a home, you are home. \\
        \textbf{Ground-Truth (Baseline):} People commit suicide and violence when there is no security. \\
        \textbf{\textsc{Neon}:} A home is a place for peace, then it is not a place for violence.\\
        \textbf{Human-annotated: }Safety and Security place is our home.\\
        }\\
    \bottomrule
    \end{tabular}
    \caption{Case study of the ComVE task. More instances can be found in Appendix \ref{app:case}.}
    \label{tab:case_study}
\end{table}

\section{Conclusion and Future Work}
In this paper, we propose a two-phase framework \textsc{Neon} to help large PLMs generate explanations by implicitly identifying conflict points in the statement. In the first phase, we generate a bunch of correct instantiations with a commonality based on the false statement. In the second phase, given both generated correct instantiations and the false statement, we adopt prompts to generate explanations according to their differences. Experiments in the unsupervised setting show that our proposed framework significantly outperforms baselines in both automatic and human evaluations. Furthermore, our analysis shows the effectiveness, robustness, and generality of \textsc{Neon}. We regard \textsc{Neon} as a first attempt towards using methods based on conflict points, which we argue is an important factor in solving the explanation tasks. In the next step, we will try to incorporate conflict points into the textual representations, e.g. through contrastive learning.

\section*{Acknowledgments}
We would like to thank the anonymous reviewers for their insightful and constructive feedback. We appreciate Tao Yu for the resource of OPT-175B; Jiacheng Ye and Lei Li for their valuable discussions.
This work is supported by the Shanghai Committee of Science and Technology (Grant No. 21DZ1100100), and National Natural Science Foundation of China (No. 61925601).

\bibliography{aaai23}

\clearpage
\appendix

\section{Prompt Construction} \label{app:Prompt}

\subsection{Correct Instantiations Generation}
In the first phase, we adopt the few-shot setting to generate correct instantiations. Due to the limitation of max length, we concatenate 16 examples as prompts. The instances of ComVE and e-SNLI tasks are shown in Tables \ref{tab:incontext} and \ref{tab:incontext_app}, respectively.

\begin{table}[h!]
    \centering
    \begin{tabular}{c}
    \toprule
    \rowcolor[gray]{0.95} 
    \textbf{e-SNLI Task} \\
    \midrule
    \makecell[l{p{8cm}}]{Task: Given the premise and the incorrect statement, generate the correct statement. \\
    \\
    \textcolor{gray}{/* Example 1 */} \\
    Premise: \textsl{This church choir sings to the masses as they sing joyous songs from the book at a church.}\\
    Incorrect statement: \textit{A choir singing at a baseball game.}. \\
    Correct statement: \textbf{\textit{The church is filled with song.}} \\
    \\
    \textcolor{gray}{/* Test data */} \\
    Premise: \textsl{A woman with a green headscarf, blue shirt and a very big grin.}\\
    Incorrect statement: \textit{The woman has been shot.} \\
    Correct statement:
    } \\
    \bottomrule
    \end{tabular}
    \caption{The prompt instances of in-context learning in correct instantiations generation: presented are the \textsl{premise}, the \textit{incorrect statements} and the \textbf{\textit{correct statements}}. In practice, we use 16 examples per prompt in the first phase.}
    \label{tab:incontext_app}
\end{table}

\subsection{Unsupervised Explanation Generation}
In the second phase, we use the zero-shot setting to generate explanations for false statements. The instances of \textsc{Neon} and baselines in both ComVE and e-SNLI tasks are shown in Table \ref{tab:phase2_comve_app} and \ref{tab:phase2_esnli_app}. The \texttt{[Guided Hints]} in different methods can be replaced by its corresponding statements, e.g. generated instantiations in \textsc{Neon}.

\begin{table}[h!]
    \centering
    \begin{tabular}{l|p{5.5cm}}
    \toprule
    \rowcolor[gray]{0.95} 
    \multicolumn{2}{c}{\textbf{ComVE Task}} \\
    \midrule
    \textbf{Origin} & \textit{John put an elephant into the fridge}. This statement is wrong because: \\
    \midrule
    \textbf{Random} & \multirow{5}{5.5cm}{Given the facts: \texttt{[Guided Hints]}, Explain the following statement based on its difference with the facts: \textit{John put an elephant into the fridge}. The explanation is:} \\
    \textbf{Retrieval} & \\
    \textbf{Ground-truth} \\
    \textbf{Top-1} \\
    \textbf{\textsc{Neon}} \\
    \bottomrule
    \end{tabular}
    \caption{The instances of our commonsense explanation generation phase for the ComVE task: presented are the \textit{incorrect statements} and the \textbf{\textit{correct statements}}.}
    \label{tab:phase2_comve_app}
\end{table}

\begin{table}[h!]
    \centering
    \begin{tabular}{l|p{5.5cm}}
    \toprule
    \rowcolor[gray]{0.95} 
    \multicolumn{2}{c}{\textbf{e-SNLI Task}} \\
    \midrule
    \textbf{Origin} & Based on the context that \textsl{a woman with a green headscarf, blue shirt and a very big grin}, explain why the following sentence is wrong: \textit{The woman has been shot}. The explanation is:\\
    \midrule
    \multirow{8}{*}{\makecell[l]{\textbf{Random} \\
    \textbf{Retrieval}\\
    \textbf{Ground-truth}\\
    \textbf{Top-1} \\
        \textbf{\textsc{Neon}} \\
    }}
    & Based on the fact that \textsl{a woman with a green headscarf, blue shirt and a very big grin}, it is correct that \texttt{[Guided Hints]}. Based on the fact that \textsl{a woman with a green headscarf, blue shirt and a very big grin}. However, it is wrong that \textit{The woman has been shot}. The explanation is that: \\
    \bottomrule
    \end{tabular}
    \caption{The instances of our commonsense explanation generation phase for the e-SNLI task: presented are the \textsl{premise}, the \textit{incorrect statements} and the \textbf{\textit{correct statements}}.}
    \label{tab:phase2_esnli_app}
\end{table}

\section{Constrained Text Generation} \label{app:cgmh}

In this section, we mainly introduce the details of modification actions. Following \citet{chen2021unsupervised}, we have three token-level modification actions replacement/insertion/deletion in a sampled distribution 0.7/0.2/0.1.

For \textit{replacement}, given the false statement $\boldsymbol{x} = \{x^1, \cdots, x^i, \cdots, x^n\}$, this action replaces the token $x^i$ with $x^r$ which sampled from a candidate set. Then we use a masked language model, e.g. BERT \citep{devlin2018bert} or RoBERTa \citep{liu2019roberta}, to compute the probability of the selected token $x^r$ according to the other tokens $\{\boldsymbol{x}\backslash{x^i}\}$, where the formulation is $P_\textnormal{MLM}({x^i}^* = x^r|\{\boldsymbol{x}\backslash{x^i}\})$.

As for \textit{insertion}, we firstly insert a \texttt{[MASK]} token into the false statement $\boldsymbol{x}$ to obtain a new statement $\boldsymbol{x}'$ with $n+1$ tokens, i.e. $\boldsymbol{x}' = \{x^1, \cdots, \texttt{[MASK]}, \cdots, x^n\}$. Then, we perform the replacement action discussed above on the \texttt{[MASK]} token.

Finally, the \textit{deletion} action is simple. If the sampled action is deletion, we will directly delete the token $x^i$ to obtain a new statement $\boldsymbol{x}' = \{x^1, \cdots, x^{i-1}, x^{i+1}, \cdots, x^n\}$.

\section{Effects on Model Size} \label{app:model}
\begin{figure}[h!]
    \centering
    \includegraphics[width=0.5\textwidth]{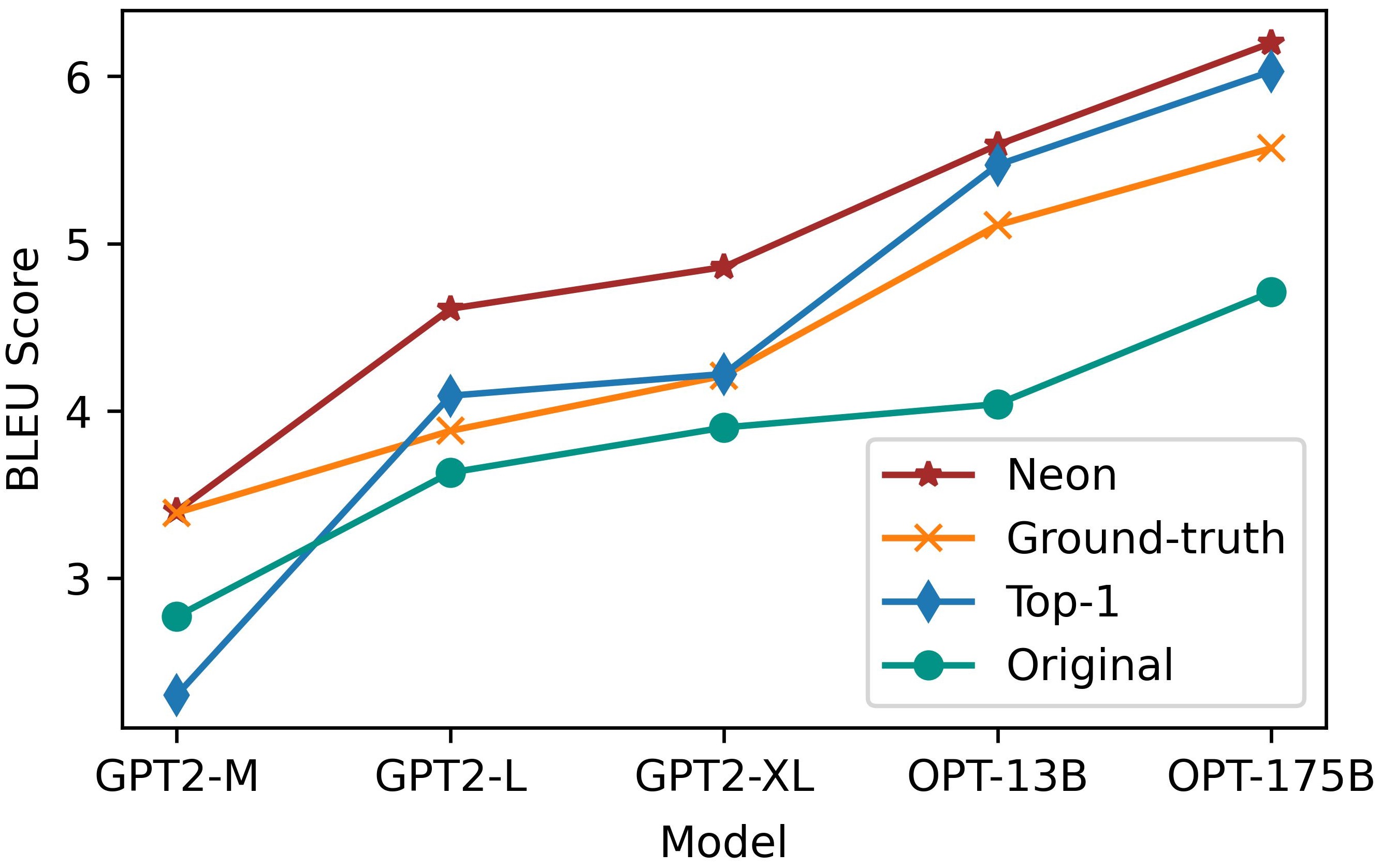}
    \caption{Model performance of increasing model scales in the e-SNLI task.}
    \label{fig:model_size_app}
\end{figure}

In this section, we detect the model performance with increasing model scales in both ComVE and e-SNLI tasks. The performances of ComVE and e-SNLI tasks are shown in Figures \ref{fig:model_size} and \ref{fig:model_size_app}, respectively. The model performances of the e-SNLI task are almost consistent with the ComVE task. \textsc{Neon} surpasses other baselines with increasing model scales.

\section{Robustness of Prompting} \label{app:robust}
In this section, we detect the model performance with different templates. The independently-written templates of ComVE and e-SNLI are shown in Table \ref{tab:templates_comve_app}.
Furthermore, their corresponding model performances are shown in Table \ref{tab:robust_app}.

\begin{table*}[t!]
    \centering
    \begin{tabular}{l|p{7cm}|p{7cm}}
    \toprule
    \textbf{Type} & \textbf{ComVE} & \textbf{e-SNLI} \\
    \midrule
    \textbf{Annotator (B)} & Given the facts: \textbf{\textit{1. John put a turkey into the fridge, 2. John put a peach into the fridge, 3. John put a bowl into the fridge}}, and the hypothesis: \textit{John put an elephant into the fridge}. The hypothesis is wrong because: & The context is \textsl{a woman with a green headscarf, blue shirt and a very big grin}. Based on the facts that \textbf{\textit{1. the woman is very happy, 2. the woman smiling. 3. the woman is happy}}, explain why the following sentence is wrong: \textit{The woman has been shot}. The explanation is:\\
    \midrule
    \textbf{Annotator (C)} & Given correct facts: \textbf{\textit{1. John put a turkey into the fridge, 2. John put a peach into the fridge, 3. John put a bowl into the fridge}}, and false statement: \textit{John put an elephant into the fridge}. Based on the difference between facts and the statement, the statement is wrong because: & Based on the facts that \textsl{a woman with a green headscarf, blue shirt and a very big grin}, it is correct that \textbf{\textit{1. the woman is very happy, 2. the woman smiling. 3. the woman is happy.}} However, it is wrong that \textit{The woman has been shot}. The explanation is that:\\
    \midrule
    \textbf{Instruction Style} & Facts: \textbf{\textit{1. John put a turkey into the fridge, 2. John put a peach into the fridge, 3. John put a bowl into the fridge}}. False statement: \textit{John put an elephant into the fridge}. Explanation: & Context: \textsl{a woman with a green headscarf, blue shirt and a very big grin}. Correct facts: \textbf{\textit{1. the woman is very happy, 2. the woman smiling. 3. the woman is happy.}}. False statement: \textit{The woman has been shot}. Explanation: \\
    \bottomrule
    \end{tabular}
    \caption{The templates of independently-written for the ComVE and e-SNLI tasks: presented are the \textsl{premise}, \textit{incorrect statements} and the \textbf{\textit{correct statements}}.}
    \label{tab:templates_comve_app}
\end{table*}

\begin{table}[htbp!]
    \centering
    \resizebox{0.5\textwidth}{28mm}{
    \begin{tabular}{ccccc}
    \toprule
        \textbf{Template} & \textbf{BLEU} & \textbf{ROUGE} & \textbf{BERTScore} & \textbf{S-BERT} \\
        \midrule
        \rowcolor[gray]{0.95} \multicolumn{5}{c}{\textbf{ComVE}} \\
        \midrule
        \textbf{Original} & 1.88 & 20.21 & 48.68 & 51.82 \\
        \textbf{\textsc{Neon}} & 3.39 & 22.50 & 51.50 & 54.52 \\
        \textbf{Annotator (B)} & 3.12 & 20.97 & 48.97 & 52.71 \\
        \textbf{Annotator (C)} & 2.93 & 21.50 & 49.60 & 52.84 \\
        \textbf{Instruction Style} & 1.98 & 19.26 & 47.97 & 47.35 \\
        \midrule
        \rowcolor[gray]{0.95} \multicolumn{5}{c}{\textbf{e-SNLI}} \\
        \midrule
        \textbf{Original} & 4.71 & 25.38 & 50.92 & 46.39 \\
        \textbf{\textsc{Neon}} & 6.20 & 27.28 & 53.87 & 51.69 \\
        \textbf{Annotator (B)} & 5.33 & 26.29 & 53.21 & 50.96 \\
        \textbf{Annotator (C)} & 6.16 & 27.43 & 54.01 & 52.08 \\
        \textbf{Instruction Style} & 5.08 & 26.58 & 51.45 & 50.92 \\
    \bottomrule
    \end{tabular}} 
    \caption{The model performances of independently-written templates to detect the robustness of \textsc{Neon}.}
    \label{tab:robust_app}
\end{table}

\section{Case Study} \label{app:case}
Finally, we present more instances of the ComVE and e-SNLI tasks in Tables \ref{tab:case_study_app}.

\begin{table*}[htbp!]
    \centering
    \begin{tabular}{p{8cm}|p{8cm}}
    \toprule
        \multicolumn{2}{c}{\textbf{Phase I: Correct Instantiations Generation}} \\
        \midrule
        \rowcolor[gray]{0.95} \multicolumn{1}{c|}{\textbf{ComVE}} & \multicolumn{1}{c}{\textbf{e-SNLI}} \\
        \midrule
        \makecell[l{p{8cm}}]{
        \textbf{False Statement:} A home is a place for violence.\\
        \textbf{Retrieval-BM25:} 1. The effect of fighting war is proliferating violence. 2. going to a restaurant requires  leaving home. 3. . 4. Trash is something that is discarded, dump is a word that means discard. 5.  A landfill is a place for trash.\\
        \textbf{Retrieval-SBERT:} 1. The effect of fighting war is proliferating violence. 2.  going to a restaurant requires  leaving home. 3. . 4. the home is the home. 5.  the home is the place you live at. \\
        \textbf{\textsc{Neon} w/ CGMH:} 1. A home is a place for everyone. 2. A home is no place for babies. 3. A home is no place for violence. 4. A home is no place for violence. 5. A home is no place for violence.  \\
        \textbf{\textsc{Neon} w/ In-context:} 1. A home is a place for safety. 2. A home is a place for love. 3. A home is where you live. 4. A home is a place for shelter. 5. A home is a place of peace.  \\
        \textbf{Human-annotated:} A home gives a person a sense of security. \\
        } & 
        \makecell[l{p{8cm}}]{
        \textbf{Premise}: A woman working long hours. \\
        \textbf{Contradiction Statement:} A woman is sleeping.\\
        \textbf{Retrieval-BM25:} 1. Another way to say "A woman is an adult female person. 2. `` is '' A female adult person is called a woman. 3. ''. 4. . 5. A French Horn is a difficult instrument to play well .\\
        \textbf{Retrieval-SBERT:} 1. a person can sleep in bed. 2. . 3. a house is for sleeping. 4. . 5. a train seat is for sleeping.\\
        \textbf{\textsc{Neon} w/ CGMH:} 1. A woman who is working in a factory. 2. This is working in a factory. 3. A woman, working in a factory. 4. This woman is working in a factory. 5. A woman, working in a factory.  \\
        \textbf{\textsc{Neon} w/ In-context:} 1. A woman is working, 2. A woman working long hours in a office; 3. A woman is working long hours; 4. A woman is working, 5. A woman works long hours.  \\
        \textbf{Human-annotated:} A woman is working. \\
        }
        \\
        \midrule
        \multicolumn{2}{c}{\textbf{Phase II: Unsupervised Explanation Generation}} \\
        \midrule
        \rowcolor[gray]{0.95} \multicolumn{1}{c|}{\textbf{ComVE}} & \multicolumn{1}{c}{\textbf{e-SNLI}} \\
        \midrule
        \makecell[l{p{8cm}}]{
        \textbf{Original:} That's the wrong definition of home. A place isn't a home, you are home. \\
        \textbf{Retrieval-BM25:} A home is not a place for violence. You don't have kids, huh?\\
        \textbf{Retrieval-SBERT:} The effects of violence are increased in a restaurant.\\
        \textbf{Ground-Truth (Baseline):} People commit suicide and violence when there is no security. \\
        \textbf{\textsc{Neon} w/ CGMH:} A home is no place for violence, right? Nope.\\
        \textbf{\textsc{Neon} w/ In-context:} A home is a place for peace, then it is not a place for violence.\\
        \textbf{Human-annotated: }Safety and Security place is our home.\\
        } & \makecell[l{p{8cm}}]{
        \textbf{Original:} Long hours can give you many different kinds of tiredness. \\
        \textbf{Retrieval-BM25:} A woman is a female person. It's A woman is an adult female person. \\
        \textbf{Retrieval-SBERT:} A woman working long hours is not sleeping. It is that she is having a rest or she is dead tired. And because she is not alive.\\
        \textbf{Ground-Truth (Baseline):} She may be enjoying her work. Then she would never be working long hours. 
        \textbf{\textsc{Neon} w/ CGMH:} sleeping in fact: present continuous.\\
        \textbf{\textsc{Neon} w/ In-context:} If a woman is sleeping, she should not be working. A woman is working long hours.\\
        \textbf{Human-annotated:} The woman is either working or sleeping.\\
        }\\
    \bottomrule
    \end{tabular}
    \caption{Case study of the ComVE and e-SNLI tasks.}
    \label{tab:case_study_app}
\end{table*}


\end{document}